\documentclass[conference]{IEEEtran}
\IEEEoverridecommandlockouts
\usepackage{cite}
\usepackage{amsmath,amssymb,amsfonts}
\usepackage{algorithmic}
\usepackage{graphicx}
\usepackage{textcomp}
\usepackage{xcolor}
\usepackage{dirtytalk}

\usepackage[inline]{enumitem}
\usepackage[breaklinks]{hyperref}
\usepackage[section]{placeins}

\def\BibTeX{{\rm B\kern-.05em{\sc i\kern-.025em b}\kern-.08em
    T\kern-.1667em\lower.7ex\hbox{E}\kern-.125emX}}
\begin{document}

\title{Dynamic Open Vocabulary Enhanced Safe-landing with Intelligence (DOVESEI)}

\author{
        Haechan Mark Bong$^{*}$,
	Rongge Zhang$^{*}$,
	Ricardo de Azambuja$^{*}$,
        Giovanni Beltrame$^{*}$
	\thanks{$^{*}$MISTLab, \'Ecole Polytechnique Montr\'eal, Montr\'eal, Canada}
	\thanks{{\tt Contact: giovanni.beltrame@polymtl.ca}}
        \thanks{This work was supported by the National Research Council Canada (NRC).}
}

\maketitle

\begin{abstract}
Drawing inspiration from the UNIX culture of \say{Do One Thing And Do It Well}, this work targets what we consider to be the foundational step for urban airborne robots, a safe landing. Furthermore, our attention is directed toward what we deem the most crucial aspect of the safe landing perception stack: segmentation. In order to tackle the challenge of ensuring safe landings, we present a streamlined reactive UAV system that employs visual servoing by harnessing the capabilities of contemporary open vocabulary image segmentation. This proposed approach can adapt to various scenarios with minimal adjustments to parameters, bypassing the necessity for extensive data accumulation for refining internal models, thanks to its open vocabulary methodology. Given the limitations imposed by local authorities (e.g. altitude, noise levels), our primary focus centers on operations originating from altitudes of 100 meters. This choice is deliberate, as numerous preceding works have dealt with altitudes up to 30 meters, aligning with the capabilities of small stereo cameras. Consequently, we leave the remaining 20m to be navigated using conventional 3D path planning methods.

Utilizing monocular cameras and image segmentation, our findings demonstrate the system's capability to successfully execute landing maneuvers at altitudes as low as 20 meters, all the while ensuring collision avoidance. However, this approach is vulnerable to intermittent and occasionally abrupt fluctuations in the terrain segmentation between frames in a video stream. To address this challenge, we enhance the image segmentation output by introducing what we call a dynamic focus: a masking mechanism that self adjusts according to the current landing stage. This dynamic focus guides the control system to avoid regions beyond the drone's safety radius projected onto the ground, thus mitigating the problems with fluctuations. Through the implementation of this supplementary layer, our experiments have reached improvements in the landing success rate of almost tenfold when compared to global segmentation. 
All the source code is open source and available online.~\footnote{
\hyperlink{https://github.com/MISTLab/DOVESEI}{https://github.com/MISTLab/DOVESEI}}

\end{abstract}


\section{Introduction}

Logistics stand as a pivotal element across diverse sectors, ranging from e-commerce operations to complex military undertakings. 
Application of Uncrewed~\cite{garber2012style, koren_2019} Aerial Vehicles (UAVs) is becoming part of research and industrial interest. Within this context, autonomous robots have emerged as an extensively sought-after solution. Notably, in modern urban environments, aerial robots are being explored as a compelling avenue to enhance last-mile delivery efficiency and reduce carbon footprint. However, challenges concerning safety have significantly hindered the widespread adoption of flying robots in more densely populated areas. When not adequately designed and operated, they can represent a possible threat to structures, vehicles and the public in general, especially if problems arise with their geolocation and other sensory information such that it could impede safe landing. Therefore, our aim is to achieve secure landings without the need for external communication, relying solely on onboard computational capabilities and perceptual abilities of compact, lightweight cameras. 

Many previous systems dealt with automatic UAV landing, but they would limit the maximum altitude~ \cite{Mittal2018VisionbasedAL,chatzikalymnios2022,7138988,rabah2018,6884813} (e.g. the OAK-D~Lite~\cite{OAKDLite2023} has a small baseline, 75mm, limiting its maximum depth range to less than 19m), use more expensive and heavier sensors~\cite{cesetti2010,rabah2018,6836176} (e.g. 3D LiDARs), or scan and create a map of the environment using Simultaneous Localization and Mapping (SLAM)~\cite{Mittal2018VisionbasedAL,chatzikalymnios2022,9981152} before deciding where to land. Finally, past use of machine learning models meant those systems would not tolerate domain shifts as they were trained and tested for very specific scenarios. 
Safe landing zone detection through segmentation needs to work in different scenarios. We proposed to use an \say{open vocabulary} based model~\cite{radford2021learning,Luddecke_2022_CVPR}, which allows to fine tune any internal model by changing only its prompt, without needing extensive collection of data. As the system in emergency landing scenario may not be able to receive or send data, external positioning like a Global Navigation Satellite System (GNSS), Ultra-Wideband Positioning (UWB), laser or radio guidance, or external markers (e.g. APRIL, ArUco) will not be available. On the other hand, it is expected the UAV can be controlled by an extra onboard system through velocity commands and it is capable of rudimentary 3D position hold (e.g. barometer, range finder, optical flow or equivalent sensor) to avoid major drifting. Consequently, our method does not use odometry or a map of the environment. An inherent advantage of the proposed methodology lies in its adaptability across diverse scenarios. By requiring only minimal parameter adjustments, this approach can cater to varying environments and operational conditions without necessitating extensive data collection or recalibration.

Our motivation is to study a minimum viable system, capable of running even with only a monocular RGB camera, that can \say{dynamically focus}, by masking the received raw segmentation according to the system's current state, and leverage open vocabulary models to allow it to be easily \say{tuned} only using language without extensive data collection. Moreover, according to the onboard sensors and computational capacity, such a system can be easily used in parallel to other methods that offer odometry and 3D mapping for further improvements. Our setup is a small, lightweight, and low-power system that can be adapted to commercial drones, and it aims to find and land on a safe landing zone (SLZ) (ground level, relatively flat, grass, open field/area/park, avoiding pedestrians, vehicles, certain structures, etc.) that is accessible for recovery by its operator in case of localization (e.g. GPS) or remote-control communication failure.
In sum, we envision achieving the overarching objective of developing a compact, lightweight, onboard external controller that can be affixed to commercial UAVs, enabling them to execute safe landings even in scenarios involving internal navigational and sensory system issues, be it due to accidents, technical malfunctions, or attacks.

\section{System Design}

\subsection{Main System Architecture}
\begin{figure}[htbp]
\centerline{\includegraphics[width=1.0\linewidth]{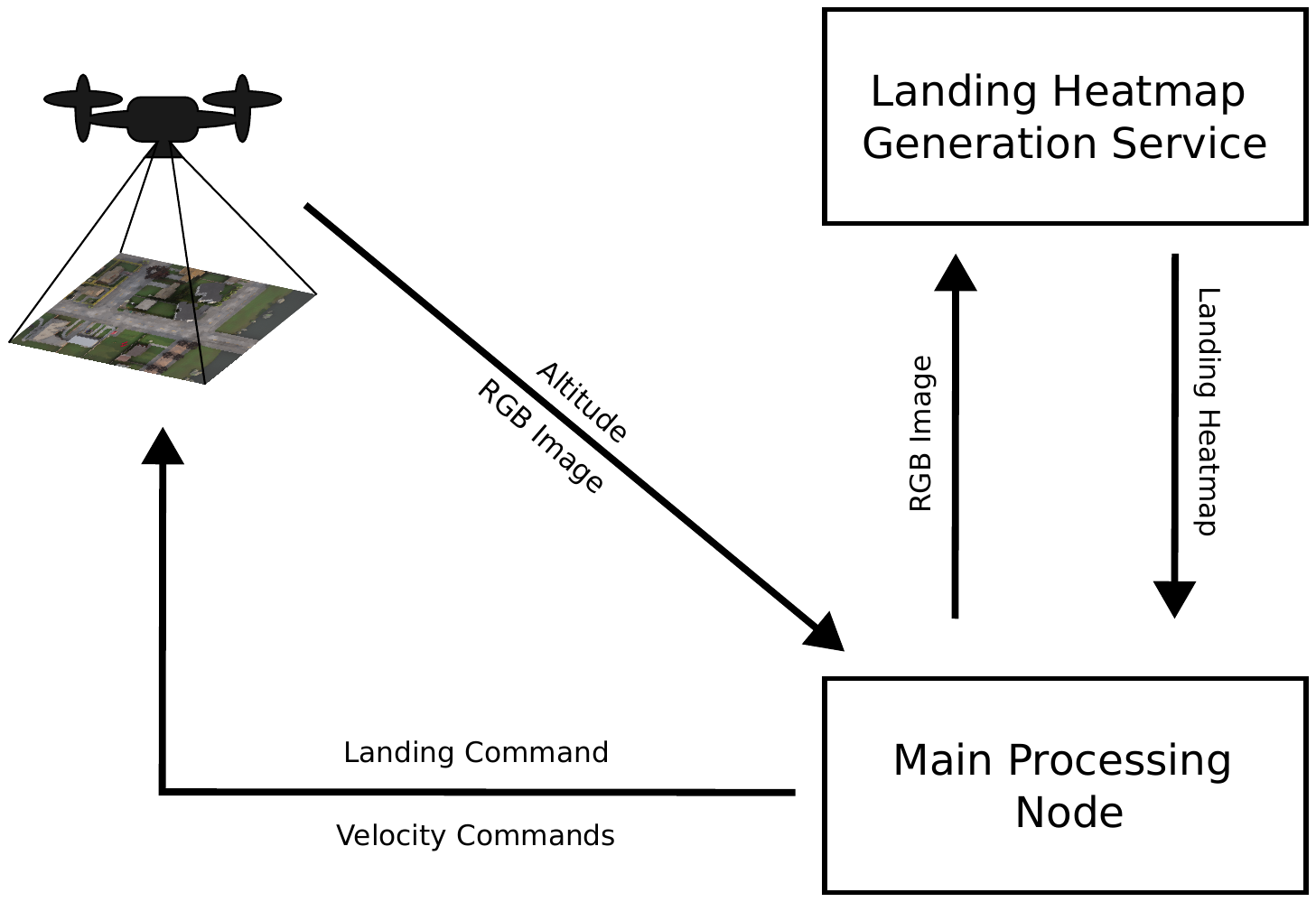}}
\caption{Our safe-landing system was implemented in ROS~2 and it is composed of three main blocks: UAV (flight controller, sensors), landing heatmap generation (receives a RGB image and produces a heatmap of the best places to land), and main processing node (orchestrates the data exchange with the UAV, sends velocity of landing commands).}
\label{methods:main}
\end{figure}
The proposed system architecture, implemented as a ROS~2~\cite{Thomas2014} package, encompasses two discrete yet interconnected processes. These processes are the Landing Heatmap Generation Service and the Main Processing Node (Fig.~\ref{methods:main}). 

\subsection{Landing Heatmap Generation Service}\label{method:heatmap}
The heatmap generator receives as inputs a RGB image and textual prompts, and it uses a pre-trained semantic segmentation model to generate (zero-shot) a comprehensive heatmap. This heatmap offers crucial insights into optimal (referred throughout this work as \say{best}) landing positions within the context of the current image frame.

The identification of these optimal landing locations (pixels) is executed through the utilization of the open vocabulary segmentation model CLIPSeg~\cite{Luddecke_2022_CVPR}, which is rooted in CLIP~\cite{radford2021learning}, a model trained on a private dataset with an estimated 400~million image-text pairs. To provide some context on CLIP's training dataset scale, the widely recognized ImageNet~\cite{imagenet2009} dataset contains \emph{only} 1.2~million images. We use CLIPSeg with a collection of words that was strategically chosen to align with the environment, such as rendered simulations and real-world satellite images. The outcomes of the segmentation process are combined to create the initial raw safe landing heatmap.

\subsection{Main Processing Node}
The main node is responsible for the high level control of the whole system and it's directly interconnected to the UAV flight controller. Its functionality can be summarised within three core components: the main state machine, raw heatmap post-processing, and dynamic focus.

\subsubsection{State Machine}\label{method:statemachine}
The state machine controls the dynamic behavior of our system and its main states are:
\begin{enumerate*}[label=\textbf{\roman*})] 
	\item Searching: coarse search for a landing spot from a safe (collision free) altitude.
	\item Aiming: refined search to better align the UAV with a safe landing spot.
        \item Landing: descend while checking for dynamic obstacles.
        \item Waiting: stop and wait if any obstacles were detected after it started landing.
        \item Climbing: climb back to the safe altitude if the waiting phase triggered a failure.
        \item Restarting: restart the coarse search by moving to a new starting position.
\end{enumerate*}

\subsubsection{Raw Heatmap Post-processing}\label{method:rawheatmap}
\begin{figure}[htbp]
\centerline{\includegraphics[width=0.85\linewidth]{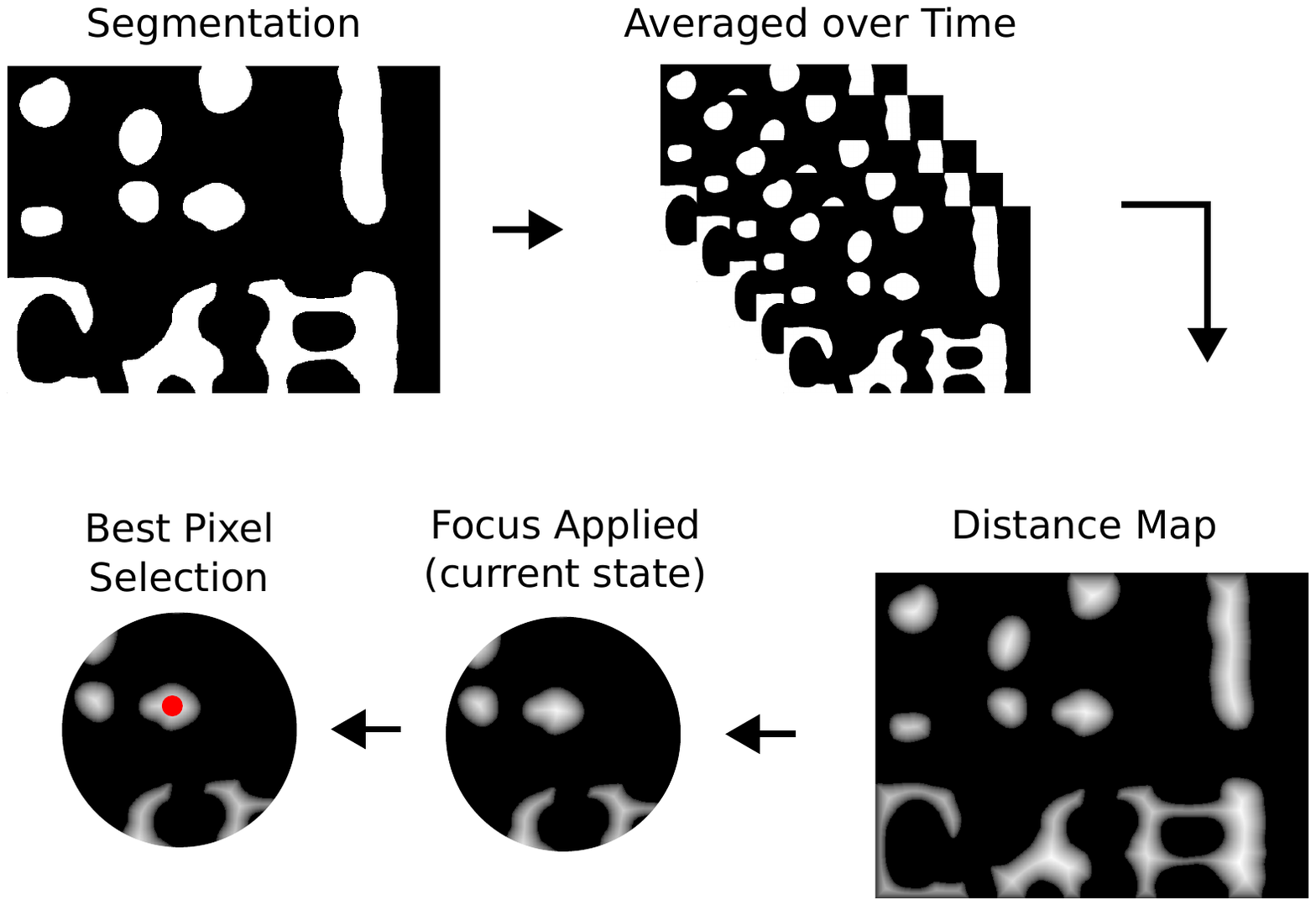}}
\caption{The visual servoing system takes as input raw segmentation heatmaps (pixels with classes considered good to land on), averages them over time (UAV's max. bank angle is limited, constraining its max. horizontal speed), creates a distance map, applies our dynamical focus masking method, and finally the objective function, Eq.~\eqref{eqn:bestpixel}, to decide on the best pixel.}
\label{methods:sementation_pipeline}
\end{figure}

The raw segmentation received from the Landing Heatmap Generation Service (\ref{method:heatmap}) is a binary mask and it alone is not enough to allow the selection of the next movement direction. To compute this direction, it's essential to choose the optimal pixel for landing, which involves passing the raw segmentation through the pipeline depicted in Fig.~\ref{methods:sementation_pipeline}. A final \say{best} pixel position is found using the objective function (the higher the value, the better) below:
\begin{equation}
P_{best} = \frac{Area}{Perimeter} \cdot {\left(C_{dist}+1\right)}^{-1}
\label{eqn:bestpixel}
\end{equation}
where $P_{best}$ is the objective function value defining the best pixel to land on (see Fig.~\ref{methods:sementation_pipeline}), $Area$ and $Perimeter$ refer to the continuous segmentation patch where the pixel is in (after the Distance Map, see Fig.~\ref{methods:sementation_pipeline}), and $C_{dist}$ is the distance to the centre of the image (the UAV's virtual position).

\subsubsection{Dynamic Focus}\label{method:dynamicfocus}
\begin{figure}[htbp]
\centerline{\includegraphics[width=1.0\linewidth]{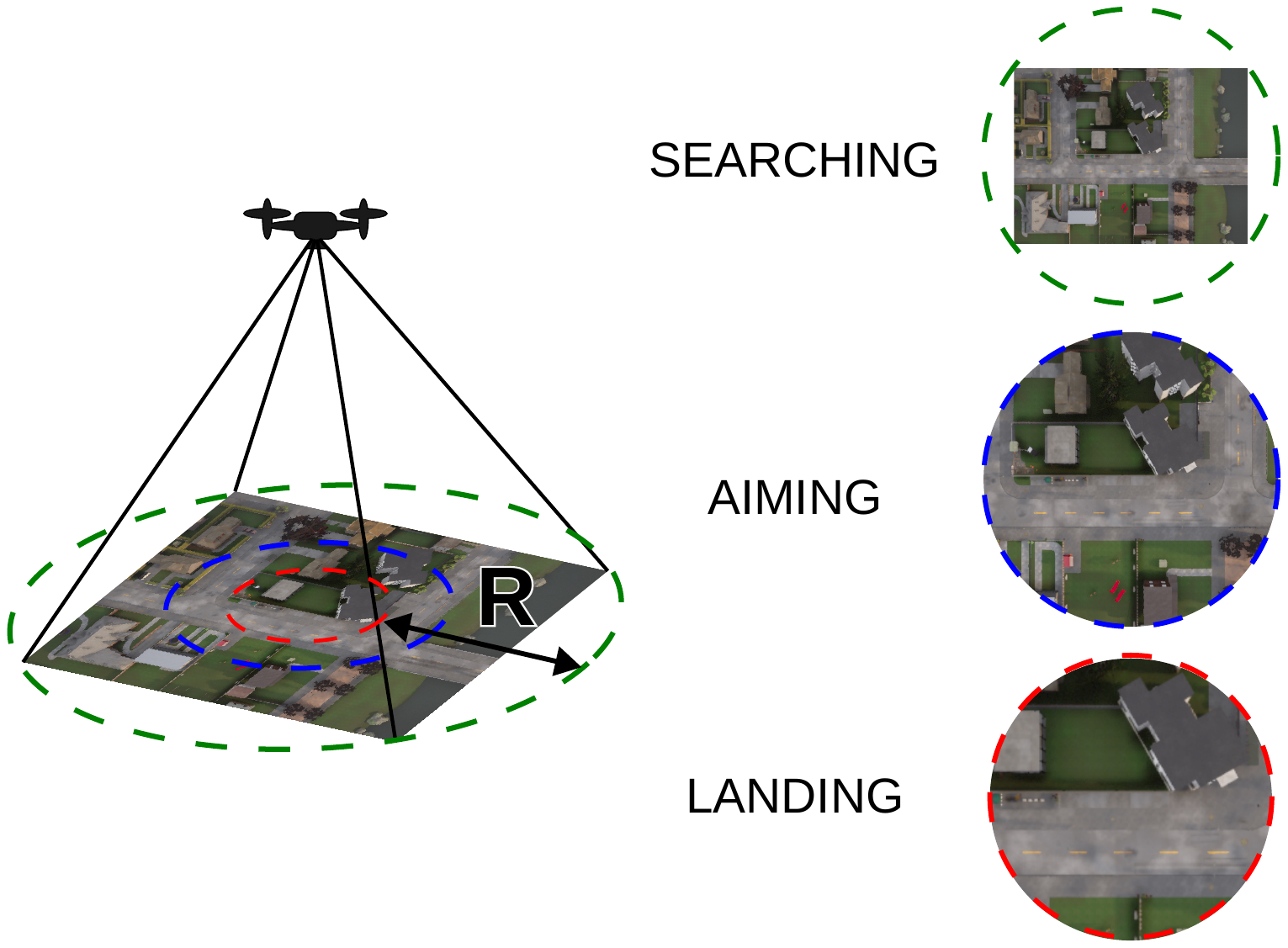}}
\caption{The focus mask radius (\textbf{R} in the illustration above) continuously varies (Eq.~\ref{eqn:dynamic_focus}), expanding or shrinking, according to the current state of the system. Its minimum size is limited by the UAV's projection on the ground (multiplier factor 6X for Aiming and 2X for Landing), while its upper limit is when the image is inscribed in the circle.}
\label{methods:focus}
\end{figure}
The dynamic focus regulates the extent to which the raw heatmap is processed by the Raw Heatmap Post-processing module (\ref{method:rawheatmap}). It \say{focuses}, i.e. applies a binary mask that covers portions of the input, on the most important areas according to the current system state (\ref{method:statemachine}) and its operation is illustrated in Fig.~\ref{methods:focus}. The dynamic (instantaneous) radius of the mask is define by the equation below:
\begin{equation}
R_{focus}(t) = R_{focus}(t - 1) + (S - R_{focus}(t - 1)\lambda
\label{eqn:dynamic_focus}
\end{equation}
where $R_{focus}$ is the dynamic focus radius (the \textbf{R} in Fig.~\ref{methods:focus}), $t$ is the current timestamp of at current state, $S$ is the UAV's safe-radius (user-defined parameter for maintaining a minimum clearance from obstacles) projection on the ground multiplied by a factor depending on the current state (6X for Aiming and 2X for Landing), and $\lambda$ is constant (empirically set to 0.1) that controls its decay or increase~speed.

\section{Experimental Setup}\label{expeimentalsetup}
\begin{figure}[htbp]
\centerline{\includegraphics[width=0.80\linewidth]{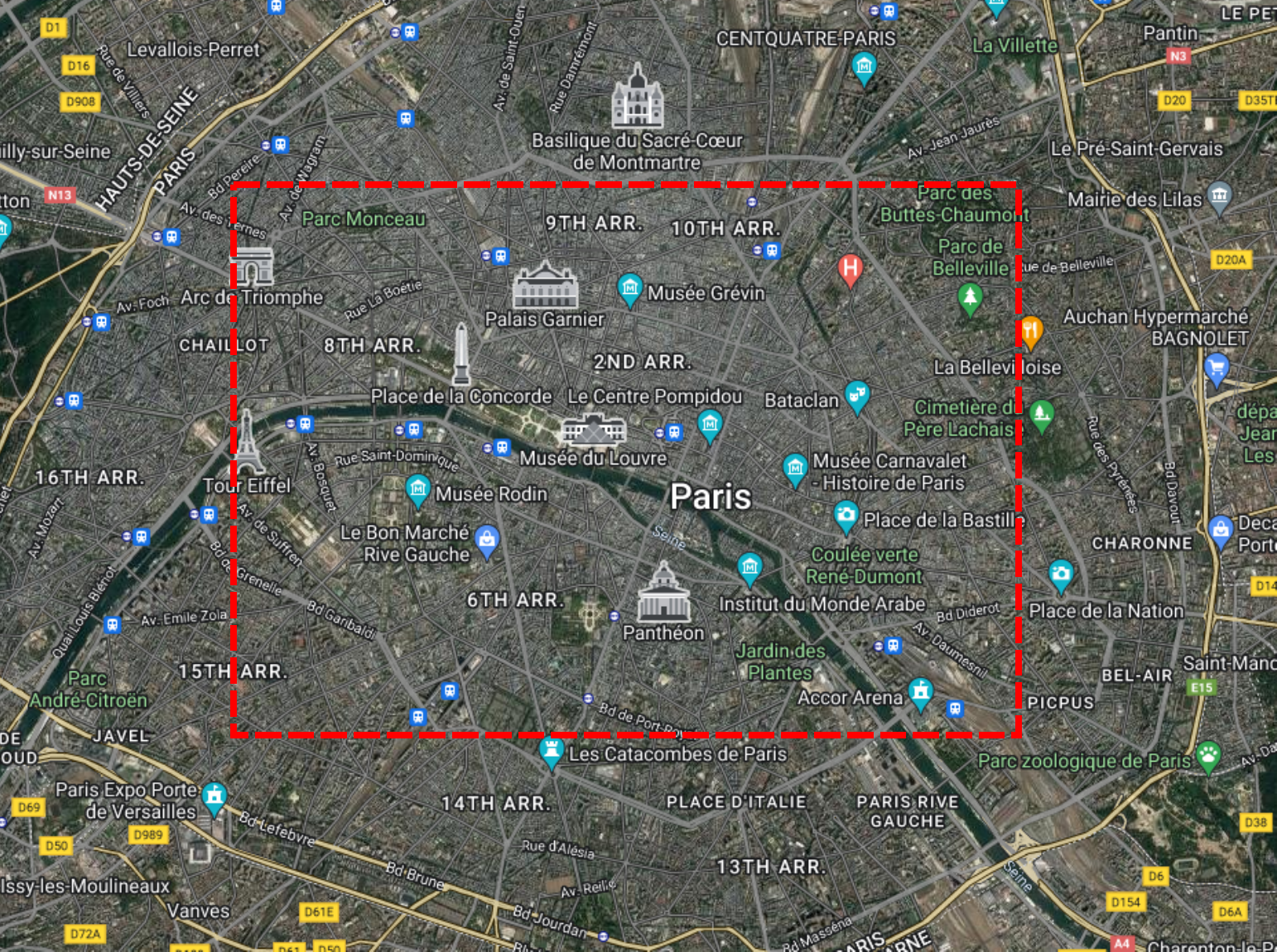}}
\caption{Satellite image of Paris showing the latitude and longitude bounding box used to uniformly sample the 50 starting positions for our experiments (red box, dashed)~\cite{GoogleMaps2023}}
\label{methods:paris_bbox}
\end{figure}

All experiments were conducted employing high-resolution satellite images from Paris, France, sourced from our open-source specialized ROS~2 package\footnote{\hyperlink{https://github.com/ricardodeazambuja/ROS2-SatelliteAerialViewSimulator}{https://github.com/ricardodeazambuja/ROS2-SatelliteAerialViewSimulator}}. We can't guarantee, but we don't expect these specific images, matched with our prompts, were ever seen by CLIP during training, therefore we consider this setting as zero-shot. A total of 50 experiments were carried out for each configuration: one with the implementation of our dynamic focus masking approach, and the other without (\emph{No Focus}). The simulated UAV always starts at an altitude of 100m and in a random location uniformly sampled from the rectangle formed by the latitude and longitude coordinates pairs (48.8810503932738, 2.2926285019834536) and (48.837230391660526, 2.389960496086875). This bounding box is illustrated in Fig.~\ref{methods:paris_bbox}.

\section{Results}

To validate our initial hypothesis that the incorporation of the Dynamic Focus~(\ref{method:dynamicfocus}) into the open vocabulary visual servoing system for safe landings was effective, we conducted tests on both systems, labeled as \emph{Dynamic Focus} and \emph{No Focus}, utilizing identical randomly sampled latitude and longitude starting positions~(\ref{expeimentalsetup}). Therefore, besides the use or not of the Dynamic Focus, all other parameters were kept the same.

\begin{table}[htbp]
\centering
\caption{Aggregated results from our landing experiments}
\begin{tabular}{|l|c|c|}
\hline
\multicolumn{1}{|c|}{}          & \textbf{Dynamic Focus} & \textbf{No Focus} \\ \hline
\textbf{Total Successful Runs}           & 29            & 3        \\ \hline
\textbf{Average Horizontal Distance (m)} & 74.40         & 81.77    \\ \hline
\textbf{Average Time Spent (s)}          & 843.98        & 943.43   \\ \hline
\end{tabular}
\label{results:aggregated_results}
\end{table}

As outlined earlier, our success criteria involve achieving an altitude of 20m over a suitable landing area, thus enabling low-cost, lightweight stereo camera and conventional 3D path planning techniques to be applied. Hence, goal positions aren't set but instead manually determined through visual inspection. Experiments exceeding the maximum allowed time (1200s) are terminated.

\begin{figure*}[htbp]
\centerline{\includegraphics[width=0.80\linewidth]{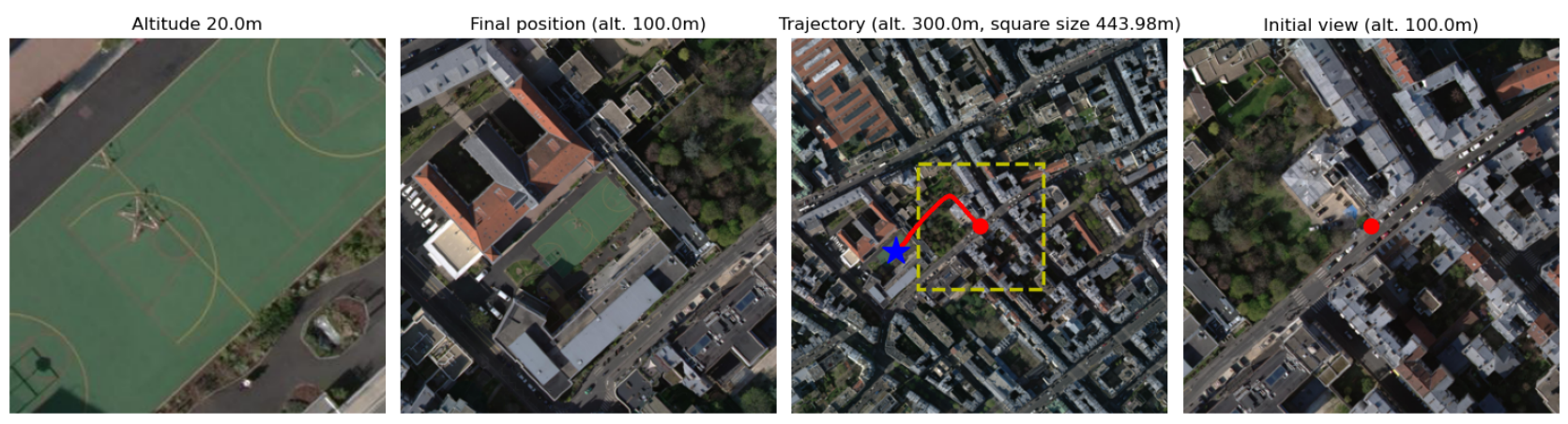}}
\caption{Example of a successful landing approach (from lat. / lon. 48.83948619062335 / 2.296169442158885 to 48.83922328688019 / 2.2948090593103774, red circle to blue star). From right to left: initial UAV's view (alt. 100m), zoom out with alt. 300m (trajectory in red, yellow dashes initial view), final location  (alt. 100m) and final UAV's view.}
\label{results:success_focus}
\end{figure*}

From a total of 50 experiments for each configuration, the setup using Dynamic Focus reached the defined safe-landing goal in 29 flights versus 3 flights without it. We also measured the total time and horizontal distance travelled in the successful trials where Dynamic Focus was, on average, 10.6\% faster and travelled a distance 9.0\% smaller~(Table~\ref{results:aggregated_results}).

\begin{figure}[htbp]
\centerline{\includegraphics[width=0.85\linewidth]{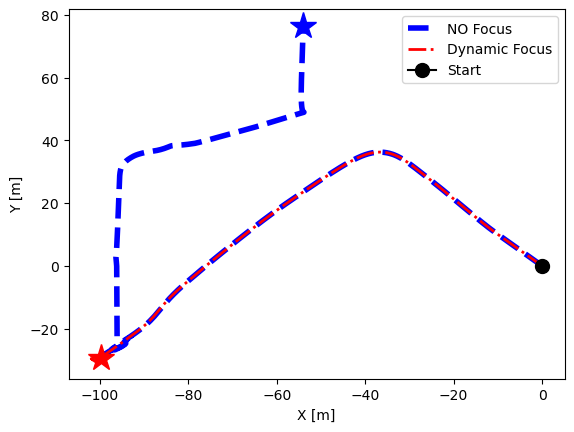}}
\caption{Paths generated for landing experiments starting at the same location as presented in Fig.~\ref{results:success_focus}, where stars indicate where each trajectory reached at the end of the experiment. Without the dynamic focus, the (global) semantic segmentation noise pushed the system away from a successful landing spot (dashed line, blue).}
\label{results:paths_experiment_046}
\end{figure}

For a clearer depiction of a typical successful run and due to space constraints in this publication, Fig.~\ref{results:success_focus} displays the satellite images alongside the UAV's path for one run using the dynamic focus. The paths for both configurations, and the same starting location, is presented in the Fig.~\ref{results:paths_experiment_046}. While the UAV is navigating only in the Searching state(~\ref{method:statemachine}), both configurations present the same behaviour, as expected. The dynamic focus allows the system to ignore \say{distractions} in the global semantic segmentation after it has found a good landing place candidate.

% \FloatBarrier
% \clearpage

\section{Discussions and Conclusions}
The primary motivation behind this work was to explore an alternative approach. Rather than fully relying on conventional solutions rooted in 3D SLAM and path planning, this study aimed on investigating the feasibility of transferring much of the system's intelligence to an open-vocabulary semantic segmentation model that uses only a RGB monocular camera, thus leaving only the final metres to be handled by a small stereo camera and traditional 3D path planing. The reliance on semantic segmentation and the observed susceptibility of such systems to fluctuations even with slight alterations in input images~\cite{azulay2019deep} prompted the development of our dynamic focus masking mechanism.

While relatively straightforward to be implemented in code, our dynamic focus functions as a selective time-varying mask significantly enhance the UAV's ability to select suitable landing sites for safe descents. Although the number of successful runs for the \say{no focus} setup was limited, with only 3 out of 50 runs, it's important to acknowledge that the inclusion of our dynamic focus yielded a substantial increase in successful outcomes, with 29 successes out of 50 runs. Although the sample size for successful runs in the \say{no focus} setup might not be statistically significant, the impact of introducing the dynamic focus seems evident in the improved success rate.

The fact that most previous works focused on low initial altitudes~\cite{Mittal2018VisionbasedAL,chatzikalymnios2022,7138988,rabah2018,6884813} or equipment with higher complexity~\cite{rabah2018,6836176} and weight that would not fit in a small UAV~\cite{cesetti2010}, makes us confident in the potential of our approach, underscoring the significance of further exploration and expansion of this research in the future. 

\section{Future Works}
Since the objective of our work is safe landing, robustness on segmentation is paramount. Such systems achieve high scores in zero-shot tasks and, therefore, offer improved generalisation capabilities over models that use a closed vocabulary (e.g. a model trained only on MS COCO~\cite{lin2014microsoft} classes). More efforts are needed to advance prompt engineering optimized for aerial images together with the understanding of what is considered safe given an environment.

Furthermore, our simulated UAV stopped at an altitude of 20m without exploring the final metres where small stereo vision sensors would have enough range to allow the use of 3D based path planning.

Such efforts would be advantageous in the context of detecting secure landing zones.

\bibliography{iros2023_focus_landing}
\bibliographystyle{IEEEtran}

\end{document}